\documentclass[11pt,a4paper]{article}
\usepackage[hyperref]{acl2021}
\usepackage{times}
\usepackage{latexsym}
\usepackage{bbm}

\usepackage{microtype}

\aclfinalcopy 
\def\aclpaperid{97} 

\usepackage{amsmath,amssymb,mathtools,bm,etoolbox}
\usepackage{graphicx}
\usepackage{booktabs}
\usepackage{multirow} 
\usepackage{color}
\usepackage{amsmath}
\usepackage{setspace}
\usepackage{diagbox}
\usepackage{makecell}
\usepackage{float}

\newcommand\blfootnote[1]{%
  \begingroup
  \renewcommand\thefootnote{}\footnote{#1}%
  \addtocounter{footnote}{-1}%
  \endgroup
}

\usepackage{algorithmicx, algorithm}
\usepackage[]{algpseudocode}

\usepackage{xcolor}

\title{Inter-GPS: Interpretable Geometry Problem Solving with Formal Language and Symbolic Reasoning}

\author{Pan Lu$^{1*}$, Ran Gong$^{1*}$, Shibiao Jiang$^{2*}$, Liang Qiu$^1$, Siyuan Huang$^1$, \\ \textbf{Xiaodan Liang$^3$, Song-Chun Zhu$^{1,4,5,6}$}\\
  $^1$Center for Vision, Cognition, Learning and Autonomy, UCLA \\
  $^2$College of Computer Science and Technology, Zhejiang University \\
  $^3$School of Intelligent Systems Engineering, Sun Yat-sen University\\
  $^{4}$Beijing Institute of General Artificial Intelligence, $^{5}$Peking University, $^{6}$Tsinghua University\\
  \texttt{panlu@cs.ucla.edu},
  \texttt{\text{\{}nikepupu, liangqiu, huangsiyuan\text{\}}@ucla.edu}, \\
  \texttt{\text{\{}jiangshibiao1999, xdliang328\text{\}}@gmail.com}, 
  \texttt{sczhu@stat.ucla.edu}
  }
  
\date{}

\begin{document}
\maketitle

\blfootnote{$^*$Equal contribution.}

\begin{abstract}
Geometry problem solving has attracted much attention in the NLP community recently. The task is challenging as it requires abstract problem understanding and symbolic reasoning with axiomatic knowledge. However, current datasets are either small in scale or not publicly available. Thus, we construct a new large-scale benchmark, Geometry3K, consisting of 3,002 geometry problems with dense annotation in formal language. We further propose a novel geometry solving approach with formal language and symbolic reasoning, called \textit{Interpretable Geometry Problem Solver} (Inter-GPS). Inter-GPS first parses the problem text and diagram into formal language automatically via rule-based text parsing and neural object detecting, respectively. Unlike implicit learning in existing methods, Inter-GPS incorporates theorem knowledge as conditional rules and performs symbolic reasoning step by step. Also, a theorem predictor is designed to infer the theorem application sequence fed to the symbolic solver for the more efficient and reasonable searching path. Extensive experiments on the Geometry3K and GEOS datasets demonstrate that Inter-GPS achieves significant improvements over existing methods.
\footnote{The project with code and data is available at \url{https://lupantech.github.io/inter-gps}.}

\end{abstract}


\section{Introduction}

\begin{figure}[t]
    \centering 
    \includegraphics[width= 0.98\linewidth]{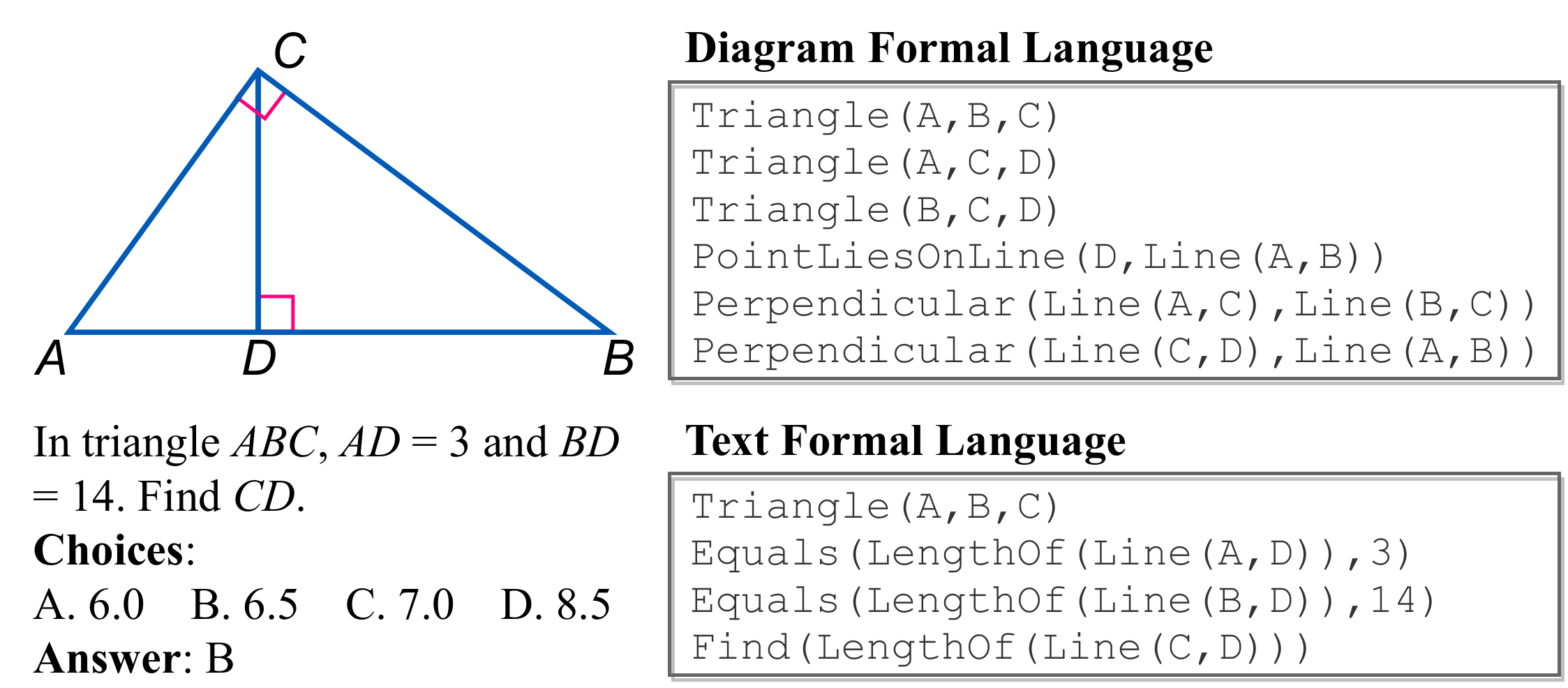}
    \vspace{0mm}
    \caption{A data example in Geometry3K dataset. Each data is annotated with formal language descriptions.}
    \vspace{0mm}
    \label{fig1:example}
\end{figure}

Geometry problem solving is a long-standing challenging task in artificial intelligence and has been gaining more attention in the NLP community recently \cite{seo2014diagram,hopkins2019semeval,sachan2020discourse}. Solving geometry problems is an essential subject in high-school education for the development of students' abstract thinking. As an example shown in Figure \ref{fig1:example}, given problem text in natural language and a corresponding diagram, one needs to identify the geometric relations, apply theorem knowledge, and conduct algebraic calculations to derive the numerical value of the answer.

Psychologists and educators believe that solving geometric problems requires high-level thinking abilities of symbolic abstraction and logical reasoning \cite{chinnappan1998schemas,nur2017geometry}. However, if algorithms take the raw problem content, it might encounter challenges to understand the abstract semantics and perform human-like cognitive reasoning for inferring the answer in the geometry domain. A formal language is composed of words from a well-formed alphabet based on a specific set of rules and is commonly used in the fields of linguistics and mathematics. Therefore, our proposed geometry solver parses the problem inputs into formal language descriptions (see examples in Figure \ref{fig1:example}) before solving the problems.


To translate the problem text and diagrams to formal descriptions, existing methods \citep{seo2015solving,sachan2017textbooks,sachan2017learning} highly depend on human annotations like symbols in diagrams as the intermediate results. Also, these methods fail to provide the explicit reasoning processes when predicting the answer. For example, \cite{seo2015solving} simplifies the problem solving task to an optimization problem to pick one that satisfies all constraints from choice candidates. Furthermore, most current datasets are either small in scale or not publicly available \cite{seo2015solving, sachan2017learning}, which further hinders the research of geometry problem solving.

To overcome these challenges, we first construct a new large-scale benchmark, called Geometry3K, to assess algorithms' performance of geometry problem solving. The Geometry3K dataset consists of 3,002 multi-choice problems as well as covers diverse geometric shapes and problem goals. 
In contrast with existing work, we also annotate each problem text and diagram with unified structural descriptions in formal language.

This paper further presents a novel geometry solving approach with formal language and symbolic reasoning, called \textit{Interpretable Geometry Problem Solver} (Inter-GPS). Inter-GPS (Figure \ref{fig:model}) develops an automatic parser that translates the problem text via template rules and parses diagrams by a neural object detector into formal language, respectively. In contrast to parameter learning, Inter-GPS formulates the geometry solving task as problem goal searching, and incorporates theorem knowledge as conditional rules to perform symbolic reasoning step by step. It demonstrates an interpretable way to tackle the task. Also, we design a theorem predictor to infer the possible theorem application sequence in Inter-GPS for the efficient and reasonable searching path. Extensive experiments on the Geometry3K and GEOS datasets show Inter-GPS achieves large improvements over existing methods.

Our contributions are three-fold: (1) we introduce a large-scale diverse benchmark of geometry problem
solving, Geometry3K, which is densely annotated with formal language; (2) we develop an automatic problem parser to translate the problem text and diagram into formal language; (3) we propose a novel interpretable problem solver that applies symbolic reasoning to infer the answer.

\section{Related Work}
\textbf{Datasets for Geometry Problem Solving.} 
Several datasets for geometry problems have been released in recent years. These include GEOS \cite{seo2015solving}, GEOS++ \cite{sachan2017textbooks}, GeoShader \cite{alvin2017synthesis} and GEOS-OS \cite{sachan2017learning} datasets. However, these datasets are relatively small in scale and contain limited problem types. For example, there are only 102 shaded area problems in GeoShader and 186 problems in GEOS. While GEOS++ and GEOS-OS contain more data of 1,406 and 2,235 problems, respectively, they have not been publicly available yet. Instead, our Geometry3K dataset features 3,002 SAT-style problems collected from two high-school textbooks that cover diverse graph and goal types. Besides, each problem in Geometry3K is annotated with dense descriptions in formal language (defined in Section \ref{sec:formal}), which makes it particularly suited for symbolic reasoning and interpretable problem solving. In order to promote follow-up work in the geometry domain, we release the dataset and evaluation baselines.

\textbf{Approaches for Geometry Problem Solving.} Due to the sparsity of appropriate data, most early works on automated geometry systems focus on geometry theorem proving \cite{wen1986basic,chou1996automated, yu2019framework,gan2019automatically}, problem synthesis \cite{alvin2014synthesis}, diagram parsing \cite{seo2014diagram}, as well as problem formalization \cite{gan2018automatic}. \cite{seo2015solving} attempt using computer vision and natural language processing techniques to solve geometry problems with problem understanding.
However, the system does not perform explicit reasoning with axiomatic knowledge as it reduces the task to an optimization problem to see which choice can satisfy all constraints. Some recent efforts \cite{sachan2017textbooks,sachan2020discourse} have been made to incorporate theorem knowledge into problem solving. They feed geometry axioms written as horn clause rules and declarations from the diagram and text parser into logical programs in prolog style to solve the problem. However, these methods fail to provide human-readable solving steps. And parameter learning on horn clause rules and built-in solvers leads to an uncontrollable search process. In contrast, our proposed Inter-GPS implements explicit symbolic reasoning to infer the answer without the help of candidate answers in an interpretable way. 

\begin{figure*}[th]
    \centering 
    \includegraphics[width= 0.91\linewidth]{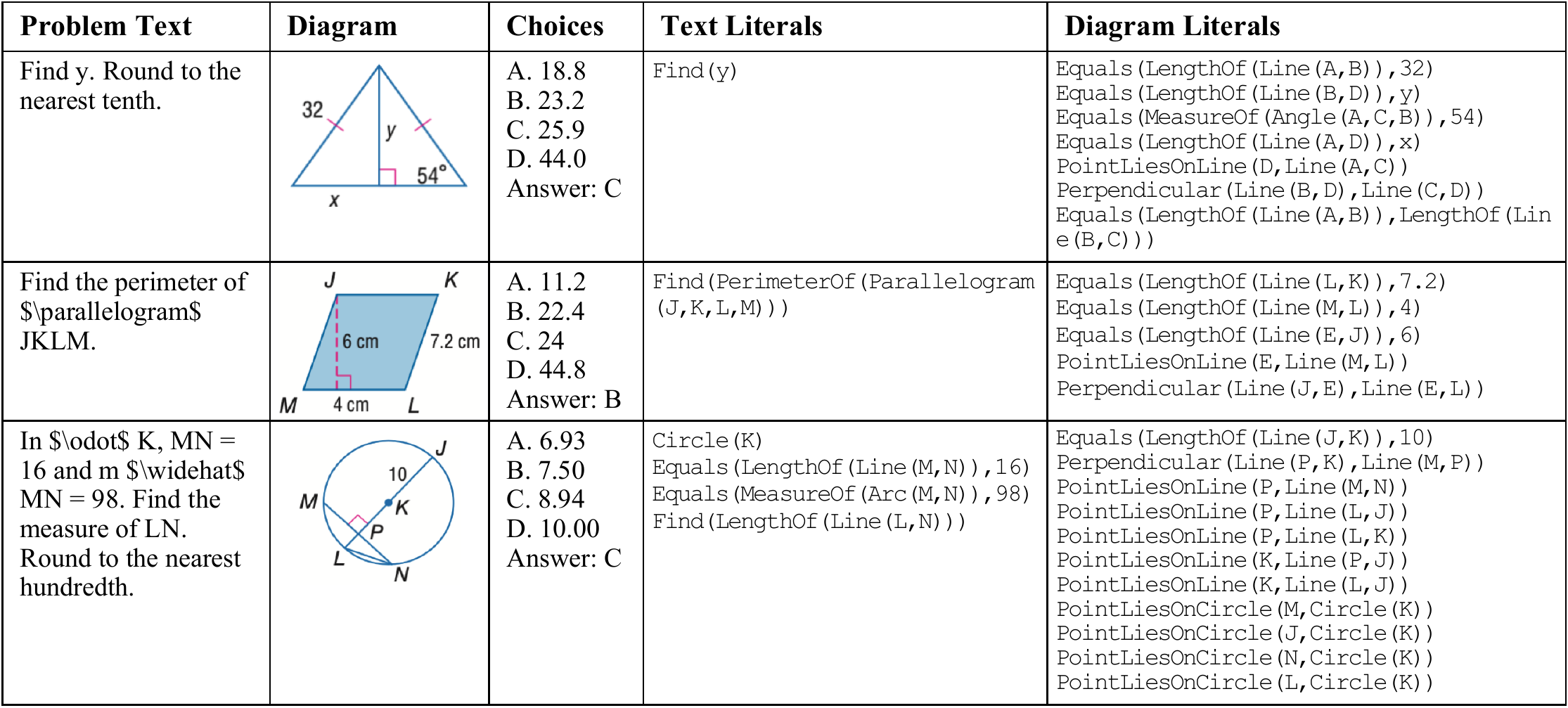}
    \caption{More data examples in the Geometry3K dataset.}
    \label{fig:data_examples}
\end{figure*}

\textbf{Interpretable Math Problem Solving.} Due to the intrinsic requirements of symbolic understanding and logical reasoning, interpretability of solvers plays an essential role in geometry problem solving. While the interpretability of geometry problem solvers is rarely explored, some pioneering work has been proposed in the general math problem solving domain. Broadly there are two main lines of achieving interpretable solving steps for math problems. The first generates intermediate structural results of equation templates \cite{huang2017learning,wang2019template}, operational programs \cite{amini2019mathqa} and expression trees \cite{wang2018translating,qin2020semantically,hong2021weakly}. The second line of work with a higher level of interpretability translates the math problems into symbolic language and conducts logical reasoning iteratively to predict the final results \cite{matsuzaki2017semantic,roy2018mapping}. Furthermore, inspired by work on semantic parsing \cite{han2005bottom,zhu2007stochastic,tu2014joint}, we claim structured diagram parsing and joint semantic representations for text and diagrams is critical in interpretable geometry problem solving.

\section{Geometry Formal Language} \label{sec:formal}

A geometry problem $P$ is defined as a tuple $(t,d,\mathbf{c})$, in which $t$ is the input text, $d$ is the diagram image and $\mathbf{c}=\{c_1, c_2, c_3, c_4\}$ is the multiple-choice candidate set in the format of numerical values. Given the text $t$ and diagram $d$, an algorithm is required to predict the correct answer $c_i \in \mathbf{c}$. We formally describe the problem in the geometric domain language $\Omega$, a set of literals composed of predicates and arguments. Basic terms used in the geometry problem solver are defined as follows.

\textbf{Definition 1.} A \textit{predicate} is a geometric shape entity, geometric relation, or arithmetic function.

\textbf{Definition 2.} A \textit{literal} is an application of one \textit{predicate} to a set of arguments like variables or constants. A set of \textit{literals} makes up the semantic description from the problem text and diagrams in the formal language space $\Omega$.

\textbf{Definition 3.} A \textit{primitive} is a basic geometric element like a point, a line segment, a circle, or an arc segment extracted from the diagram.

\begin{table}[ht]
\centering 
\footnotesize
\begin{tabular}{l|l}
	\hline
	\textbf{Terms} & \textbf{Examples} \\
	\hline
	\textit{predicate} & \footnotesize \texttt{Line}, \texttt{IntersectAt}, \texttt{IsMedianOf} \\
	\textit{literal} & \footnotesize \texttt{Find(AreaOf(Triangle(A,B,C))} \\
	\hline
\end{tabular}
\vspace{0mm}
\caption{Term examples in geometry formal language.}
\vspace{0mm}
\label{tab1:definition}
\end{table}

Table \ref{tab1:definition} lists examples of \textit{predicates} and \textit{literal} templates. There are 91 \textit{predicates} in our defined formal language, and we list them in the Tables \ref{appex:predicate1} to \ref{appex:predicate6} in the Appendix Section.


\begin{table*}[th!]
\centering 
\setlength{\tabcolsep}{3pt}
\small 
\begin{tabular}{l|ccccccc}
	\hline
	\textbf{Dataset} & \#qa & \#word & \#shape & \#goal & \#var & grade & operator type \\ 
	\hline
	GeoShader \cite{alvin2017synthesis} & 102 & / & 4 & 1 & 1 & 6-10  & \{$+$, $-$, $\times$, $\div$, $\Box^2$, $\sqrt\Box$\} \\
	GEOS \cite{seo2015solving} & 186 & 4,343 & 4 & 3 & 1 & 6-10  & \{$+$, $-$, $\times$, $\div$, $\Box^2$, $\sqrt\Box$\} \\ 
	GEOS++ \cite{sachan2017textbooks} & 1,406 & / & 4 & 3 & 1 & 6-10 & \{$+$, $-$, $\times$, $\div$, $\Box^2$, $\sqrt\Box$\} \\
	GEOS-OS \cite{sachan2017learning} & 2,235 & / & 4 & 3 & 1 & 6-10 & \{$+$, $-$, $\times$, $\div$, $\Box^2$, $\sqrt\Box$\} \\
	\textbf{Geometry3K} (ours) & 3,002 & 36,736 & 6 & 4 & 3 & 6-12 & \footnotesize \{$+$, $-$, $\times$, $\div$, $\Box^2$, $\sqrt\Box$, $\sin$, $\cos$, $\tan$\} \\
	\hline
\end{tabular}
\vspace{0mm}
\caption{Comparison of our Geometry3K dataset with existing datasets. }
\vspace{0mm}
\label{table:dataset}
\end{table*}

\section{Geometry3K Dataset}
\subsection{Dataset Collection}\label{sec:dataset}
Most existing datasets for geometry problem solving are relatively small, contain limited problem types, or not publicly available. For instance, the GEOS dataset \citep{seo2015solving} only contains 186 SAT problems. Although there are 1,406 problems in GEOS++ \citep{sachan2017textbooks}, this dataset has not been released to the public yet. Therefore, we build a new large-scale geometry problem benchmark, called Geometry3K. The data is collected from two popular textbooks for high school students across grades 6-12 by two online digital libraries (McGraw-Hill\footnote{https://www.mheducation.com/}, Geometryonline\footnote{www.geometryonline.com }). Groups of well-trained annotators with undergraduate degrees manually collect each problem with its problem text, geometry diagram, four candidate choices, and correct answer. 
In order to evaluate the fine-grained performance of geometry solvers, we label each problem data with the corresponding  problem goal and geometry shapes. 

Unlike existing datasets that only collect the problem text and diagrams, we further annotate each data in Geometry3K with dense formal language descriptions that bridge the semantic gap between the textual and visual contents as well as benefit the symbolic problem solver. The annotated formal language is used to train and evaluate our proposed problem parsers. Data examples are illustrated in Figure \ref{fig:data_examples}.


\begin{table}[t]
\centering 
\small 
\begin{tabular}{l|cccc}
	\hline
	 & Total & Train & Val & Test  \\
	\hline
	Questions & 3,002 & 2,101 & 300 & 601 \\
	Sentences & 4,284 & 2,993 & 410 & 881 \\
	Words  & 30,146 & 20,882 & 2,995 & 6,269 \\
	\hline
	Literals (Text) & 6,293 & 4,357 & 624 & 1,312 \\
	Literals (Diagram) & 27,213 & 19,843 & 2,377 & 4,993 \\
	\hline
\end{tabular}
\caption{Basic statistics of our Geometry3K dataset.}
\label{table:data_stats}
\end{table}

\subsection{Dataset Statistics}

\begin{figure}[th]
    \centering 
    \includegraphics[width= 0.9\linewidth]{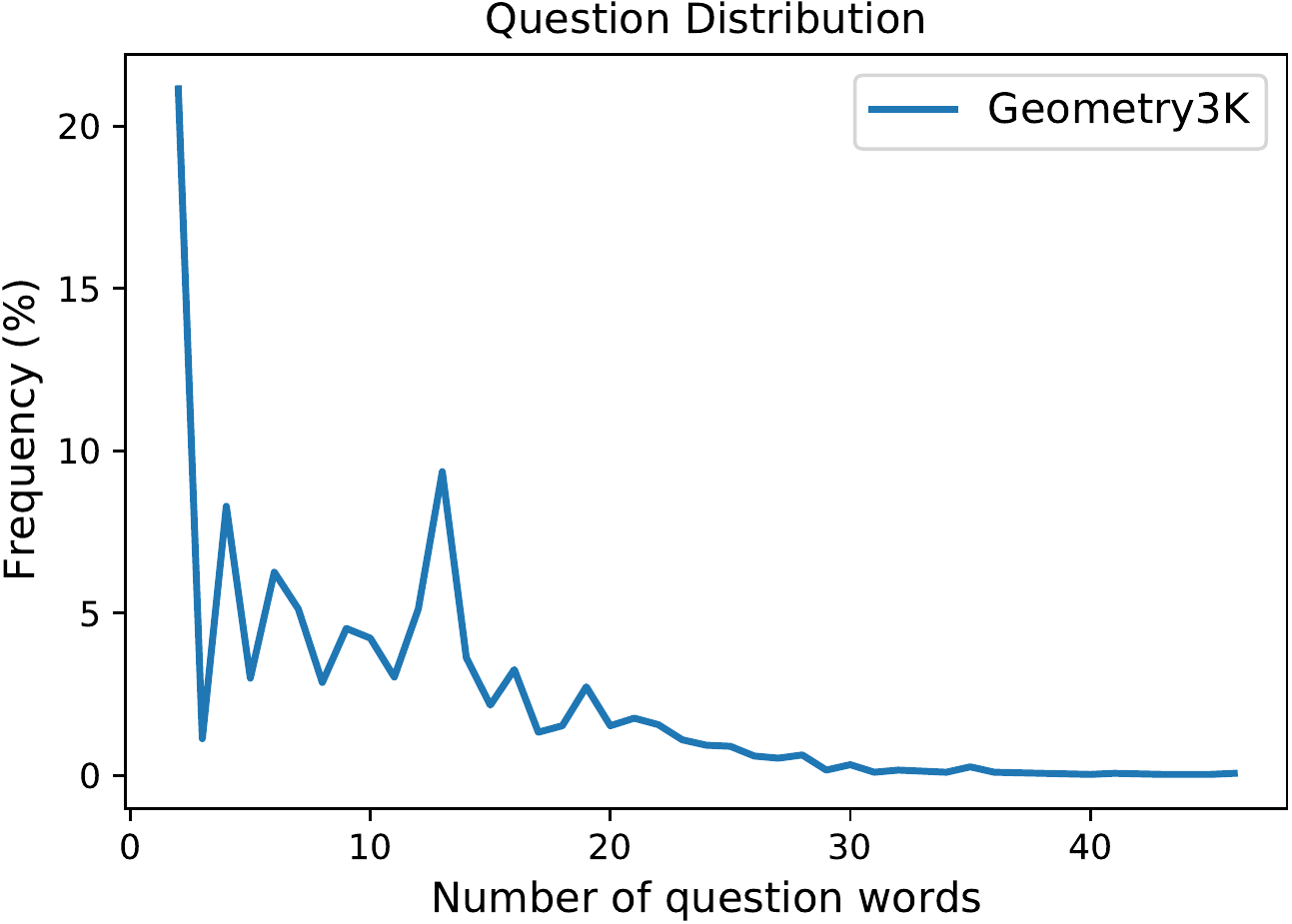}
    \caption{Question length distribution of Geometry3K.}
    \label{fig2:ques_dist}
\end{figure}

The Geometry3K dataset consists of 3,002 problems and is divided into the train, validation, and test sets with the ratio of 0.7:0.1:0.2, as shown in Table \ref{table:data_stats}. Figure \ref{fig2:ques_dist} illustrates the question distribution by the number of sentence words. The long tail in the distribution requires the geometry solvers to understand the rich semantics in the textual content.

There are 6,293 literals for the problem text and 27,213 literals for the diagrams in Geometry3K, respectively. We list the most and least frequent predicates with a frequency greater than 5 in Table \ref{table:literal}. It is shown that the predicates for the problem text are more evenly distributed than those for diagrams. This is mainly because the problem text describes diverse geometric shapes, attributes, and relations while diagrams display the basic properties of points, lines, and arcs.

\begin{table}[t!]
\centering 
\setlength{\tabcolsep}{1.5pt}
\footnotesize
\begin{tabular}{lc|lc}
	\hline
	 Predicates (Text) & \% & Predicates (Diagram) & \% \\
	\hline
    \texttt{Find} & 19.00 & \texttt{Line} & 30.89 \\
    \texttt{Line} & 14.49 & \texttt{PointLiesOnLine} & 16.66 \\
    \texttt{Equals} & 11.83 & \texttt{Equals} & 15.17 \\
    \texttt{LengthOf} & 9.53  & \texttt{MeasureOf} & 10.46 \\
    \texttt{MeasureOf} & 8.97 & \texttt{LengthOf} & 8.69  \\
    ...... &  & ...... & \\
    \texttt{CircumscribedTo} & 0.05 & \texttt{Triangle} & 0.03 \\
    \texttt{SumOf} & 0.04 & \texttt{Quadrilateral} & 0.02 \\
    \texttt{HeightOf} & 0.04 & \texttt{Kite} & 0.01 \\
    \texttt{BaseOf} & 0.04 & \texttt{HeightOf} & 0.01 \\
    \texttt{IsHypotenuseOf} & 0.04 & \texttt{Square} & 0.01 \\
	\hline
\end{tabular}
\caption{Most and least frequent predicates of formal descriptions in Geometry3K (frequency \textgreater 5).}
\label{table:literal}
\end{table}

\subsection{Comparisons with Existing Datasets}

To the best of our knowledge, currently, it is the largest geometry problem dataset. We summarize the Geometry3K dataset's main statistics and a comparison of existing datasets in Table \ref{table:dataset}. In addition to four elementary shapes (lines, triangles, regular quadrilaterals, and circles) mentioned in that GEOS dataset, Geometry3K contains irregular quadrilaterals and other polygons. Besides, in Geometry3K, there are more unknown variables and operator types that may require equation solving to find the goal of the problem. Note that 80.5\% of problems are solvable without the associated diagram in the GEOS dataset. By contrast, less than 1\% of the problems in our Geometry3K dataset could be solved when the problem diagram is not provided. In general, the statistics and comparisons above show  Geometry3K is challenging for geometry problem solvers.

\subsection{Human Performance}
As an intellectual task, it is necessary to know the human performance for geometry problems. We push the test-split data of the dataset in the crowdsourcing platform, Amazon Mechanical Turk\footnote{https://www.mturk.com/}. Each eligible annotator must have obtained a high school or higher degree and is asked to answer 10 problems in 25 minutes. To ensure annotators solving the problem to the best of their ability, they are further asked to spend at least 7 minutes on the problem set and 10 seconds on each problem. We filter out annotators who do not satisfy the requirement.  We also ask dozens of graduates majoring in science or engineering to answer these problems to evaluate human experts' performance. Table \ref{table:result} shows the human performance. Compared to random guess's accuracy of 25\%, humans achieve an overall accuracy of 56.9\%, and human experts can achieve a good performance of 90.9\%.


\begin{figure*}[th]
    \centering 
    \includegraphics[width= 0.9 \linewidth]{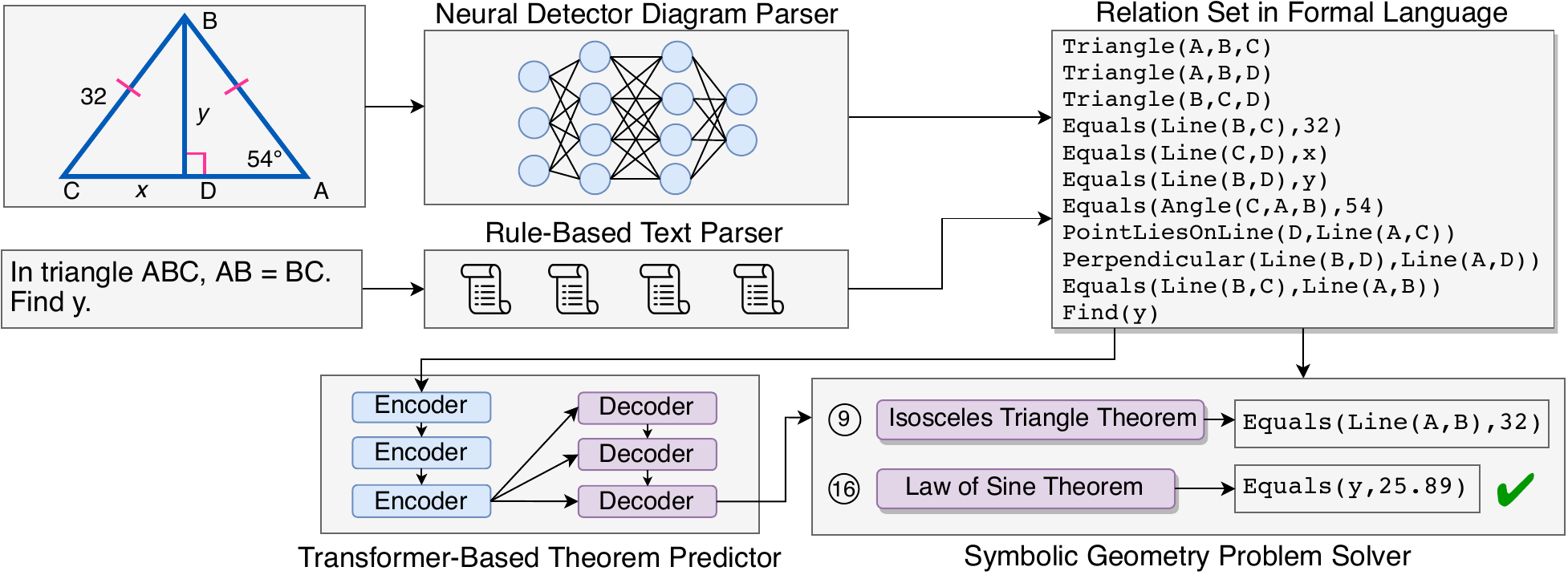}
    \caption{Given the problem diagram and text, our proposed Inter-GPS first parses the inputs into a relation set defined in formal language. Then Inter-GPS applies the theorem sequence predicted by the theorem predictor to perform symbolic reasoning over the relation set to infer the answer. \textcircled{\scriptsize{9}} and \textcircled{\scriptsize{16}} denote the theorem orders.}
    \label{fig:model}
\end{figure*}

\section{Geometry Problem Parser}
Our proposed Inter-GPS takes the problem text and diagrams as inputs and translates them into formal language descriptions automatically via the text parser (Section \ref{sec:text_parser}) and the diagram parser (Section \ref{sec:diagram_parser}), respectively.

\subsection{Text Parser} \label{sec:text_parser}
Given the word sequence of the problem text $T$, the text parser needs to translate it into a set of literals $L_t$, a sequence composed of predicates and variables. Recently, deep neural networks have achieved promising performances in sequence-to-sequence (Seq2Seq) learning tasks like machine translation \cite{sutskever2014sequence,vaswani2017attention,devlin2018bert}. However, semantic parsers using Seq2Seq learning methods are not feasible to generate satisfactory literals in the Geometry3K dataset for two reasons. Firstly, the limited scale of geometry datasets weakens these highly data-driven methods. Secondly, neural semantic parsers tend to bring noises in generated results while geometry solvers with symbolic reasoning are sensitive to such deviations. 

Inspired by previous works \citep{koo2008simple,seo2015solving,bansal2014tailoring}   that indicate the rule-based parsing method is able to obtain precise parsing results, we apply this approach with regular expressions to perform text parsing.  We also achieve a semantic text parser using BART \cite{lewis2020bart}, one of the state-of-the-art sequence learning models for comparison. 



\subsection{Diagram Parser} \label{sec:diagram_parser}
Diagrams provide complementary geometric information that is not mentioned in the problem text. Previous works \citep{seo2014diagram,seo2015solving} require manual annotations to identify symbols in the diagrams and fail to deal with special relational symbols such as \textit{parallel}, \textit{perpendicular}, and \textit{isosceles}. Instead, an automatic diagram parser without human intervention is proposed in this work and is able to detect varied diagram symbols.

The diagram parser first applies Hough Transformation \citep{shapiro2001computer} to extract geometry primitives (points, lines, arcs, and circles), following \citep{seo2015solving}. Then the diagram symbols and text regions are extracted through a strong object detector RetinaNet \cite{lin2017focal}, and the textual content is further recognized by the optical character recognition tool MathPix\footnote{https://mathpix.com/}.  After obtaining the primitive set $P$ and symbol set $S$, we need to ground each symbol with its associated primitives. \cite{seo2015solving} adapts a greedy approach where each symbol is assigned to the closest primitive without considering its validity. Instead, we formulate the grounding task as an optimization problem with the constraint of geometry relations:

\begin{equation}
\begin{aligned}
    & \min \sum_{s} dist(s_i, p_j) \times \mathbbm{1}\{ s_i \text{ assigns to } p_j \} \\
    & \text{s.t. } (s_i, p_j) \in \text{Feasibility set} ~ F,
\end{aligned} 
\end{equation}
where the $dist$ function measures the Euclidean distance between the symbol $s_i$ and primitive $p_j$. $F$ defines the geometric constraints for symbol grounding. For example, the \textit{parallel} symbol could only be assigned to two lines with the same slopes and the \textit{perpendicular} symbol is only valid to two orthogonal lines.

\section{Geometry Problem Solver}
Unlike existing methods \cite{seo2015solving,sachan2017textbooks,alvin2017synthesis,sachan2020discourse}, Inter-GPS achieves the explicit symbolic reasoning with the theorem knowledge base and the human-readable search process, shown in Figure \ref{fig:model}.


\subsection{Symbolic Geometry Solver} \label{sec:symbolic_solver}

Overall, Inter-GPS takes the relation set $\mathcal{R}$ and the theorem knowledge base set $\mathcal{KB}$ as inputs, and outputs the numeric solution $g^*$ of the problem goal $g$. The relation set $\mathcal{R}$ defines geometry attributes and relations in the given problem, and is initialized with literals from the text and diagram parsers.  $\mathcal{R}$ is further expanded with literals that are derived from definitions of geometry shapes. For example, a triangle is defined as three connected sides. So if there is a \textit{literal} \small\texttt{Triangle(A,B,C)}\normalsize, six more \textit{literals} (\small \texttt{Ponit(A)}, \texttt{Ponit(B)}, \texttt{Ponit(C)}, \texttt{Line(A,B)}, \texttt{Line(B,C)}, \texttt{Line(C,A)}\normalsize) will be appended to $\mathcal{R}$.

The theorem set $\mathcal{KB}$ is represented as a set of theorems, where each theorem $k_i$ is written as a conditional rule with a premise $p$ and a conclusion $q$. For the search step $t$, if the premise $p$ of $k_i$ matches the current relation set $\mathcal{R}_{t-1}$, the relation set is updated according to the conclusion $q$:
\begin{equation}
\begin{split}
    \mathcal{R}_{t} \gets k_i \land \mathcal{R}_{t-1}, k_i \in \mathcal{KB}.
\end{split}
\end{equation}
After the application of several theorems, equations between the known values and the unknown problem goal $g$ are established, and $g$ could be solved after solving these equations:
\begin{equation}
\begin{split}
    g^* \gets \text{\scshape SolveEquation}(\mathcal{R}_t, g).
\end{split}
\end{equation}

\subsection{Theorem Predictor (TP)}
As the geometry problems in Geometry3K are collected from high school textbooks, it might need to apply multiple theorems before the problems are solved. Intuitively, one possible search strategy is to use brute force to enumerate candidates in the theorem set randomly. The random search strategy is inefficient and might lead to problems unsolvable as there might be applications of complicated theorems in the early stage. Therefore, an ideal geometry problem solver can solve the problems using reasonable theorem application sequences. Students with good academic performance can solve a problem with prior knowledge learning from a certain amount of problem solving training. Inspired by this phenomenon, a theorem predictor is proposed to infer the possible theorem application sequence for inference after multiple attempts on the train data. Recent studies \cite{loos2017deep,balunovic2018learning} also suggest that neural guided search can speed up the search process.

There are no annotated theorem application sequences for data in Geometry3K due to tremendous worker labor. 
Thus, we randomly sample from the theorem set multiple times to generate the application sequences.
A generated sequence is regarded as positive if the geometry solver Inter-GPS solves the problem after the application of that sequence. A positive sequence with the minimum length for a problem is seen as pseudo-optimal. Finally, after attempts, we collect 1,501 training samples with the problem and its pseudo-optimal theorem application sequence. 

Given the problem formal description $L = \{l_1,...,l_m\}$, the theorem predictor aims to reconstruct the pseudo-optimal theorem  sequence $T = \{t_1,...,t_n\}$ token by token. We formulate the generation task as a sequence-to-sequence (Seq2Seq) problem and use a transformer-based model \cite{lewis2020bart} to generate theorem sequence tokens.  Specifically, the transformer decoder predicts the next theorem order $t_i$ given $T = \{t_1,...,t_i\}$. The Seq2Seq model is trained to optimize the negative log-likelihood loss:
\begin{equation}
\begin{split}
    \mathcal{L}_{\mathrm{TP}} = 
    -\sum_{i=1}^{n} 
    \log p_{\mathrm{TP}}\left(t_{i} \mid t_{1}, \ldots, t_{i-1}\right),
\end{split}
\end{equation}
where $p_{\mathrm{TP}}$ is the parametrized conditional  distribution in the theorem predictor model.

\begin{algorithm}[t]
\caption{Symbolic Geometry Solver}
\label{alg:solver}
\small
\hspace*{0.02in} {\bf Input} 
 Literals $\mathcal{L}$, goal $g$, knowledge bases \hspace*{0.0in}  $\mathcal{KB}_1$, $\mathcal{KB}_2$ \\
\hspace*{0.02in} {\bf Output}
Numeric goal value $g^*$ and theorem application $\mathcal{S}$
\begin{algorithmic}[1]
    \Function{Search}{$\mathcal{L}$, $g$, $\mathcal{KB}_1$, $\mathcal{KB}_2$}
        \State Initialize relation set $\mathcal{R}_0$ with $\mathcal{L}$, $g^* = \emptyset$, $\mathcal{S} =  \emptyset$
        \State $\mathcal{KB}_p \gets \text{\scshape TheoPredictor}(\mathcal{L})$ \Comment{Predicted}
        \For{$k_i \in \mathcal{KB}_p$} 
            \State $\mathcal{R}_{t} \gets k_i \land \mathcal{R}_{t-1}$
            \State $\mathcal{S}$.\text{\scshape append}($k_i$)
        \EndFor
        \State $g^* \gets \text{\scshape SolveEquation}(\mathcal{R}_t, g)$
        \If {$g^* \neq \emptyset$}
            \State \Return $g^*$ and $S$
        \EndIf
        \While{$g^* = \emptyset$ and $\mathcal{R}_t$ is updated}
        \For{$k_i \in \mathcal{KB}_1$} \Comment{Lower-order}
            \State $\mathcal{R}_{t} \gets k_i \land \mathcal{R}_{t-1}$
            \State $\mathcal{S}$.\text{\scshape append}($k_i$)
            \State $g^* \gets \text{\scshape SolveEquation}(\mathcal{R}_t, g)$
            \If {$g^* \neq \emptyset$}
                \State \Return $g^*$ and $S$
            \EndIf
        \EndFor
        \For{$k_i \in \mathcal{KB}_2$} \Comment{Higher-order}
            \State $\mathcal{R}_{t} \gets k_i \land \mathcal{R}_{t-1}$
            \State $\mathcal{S}$.\text{\scshape append}($k_i$)
            \State $g^* \gets \text{\scshape SolveEquation}(\mathcal{R}_t, g)$
            \If {$g^* \neq \emptyset$}
                \State \Return $g^*$ and $S$
            \EndIf
        \EndFor
        \EndWhile\label{euclidendwhile}
    \EndFunction
\end{algorithmic}
\end{algorithm}

\subsection{Low-first Search Strategy}

After the application of the theorem sequence predicted by the theorem predictor, it is likely that Inter-GPS still could not find the problem goal. Generally, humans incline to use simple theorems first when solving math problems to reduce complex calculations. If simple theorems are not tangible, they will turn to more complex theorems. On account of that, we apply an efficient search strategy with heuristics driven by subject knowledge. We categorize theorems into two groups: \textbf{lower-order} theorem set $\mathcal{KB}_1$ and \textbf{higher-order} theorem set $\mathcal{KB}_2$. The lower-order set $\mathcal{KB}_1$ (e.g, \textit{Triangle Angle-Sum Theorem}, \textit{Congruent Triangle Theorem}) only involves in two simple operations of addition and subtraction, while $\mathcal{KB}_2$ (e.g, \textit{Law of 
Sines}) requires complex calculations. In each following search step after using predicted theorems, we first enumerate theorems in the lower-order set $\mathcal{KB}_1$ to update the relation set $\mathcal{R}$:
\begin{equation}
\begin{split}
    \mathcal{R}_{t} \gets k_i \land \mathcal{R}_{t-1}, k_i \in \mathcal{KB}_1.
\end{split}
\end{equation}
If lower-order theorems fail to update $\mathcal{R}$ anymore, higher-order theorems are considered to update $\mathcal{R}$:
\begin{equation}
\begin{split}
    \mathcal{R}_{t} \gets k_i \land \mathcal{R}_{t-1}, k_i \in \mathcal{KB}_2.
\end{split}
\end{equation}


The search process stops once we find the problem goal $g$ or the search steps reach the maximum steps allowed. The whole search algorithm for Inter-GPS is presented in Algorithm \ref{alg:solver}.

\section{Experiments}
\begin{table*}[t!]
\centering 
\footnotesize
\begin{tabular}{lc|cccc|ccccc}
	\hline
	\textbf{Method} & All & Angle & Length & Area & Ratio  & Line & Triangle & Quad & Circle & Other \\
	\hline
	Random &25.0 &25.0 &25.0 &25.0 &25.0 &25.0 &25.0 &25.0 &25.0 &25.0  \\
	Human &56.9 &53.7 &59.3 &57.7 &42.9 &46.7 &53.8 &68.7 &61.7 &58.3 \\
	Human Expert &90.9 &89.9 &92.0 &93.9 &66.7 &95.9 &92.2 &90.5 &89.9 &92.3 \\
	\hline
	Q-only &25.3 &29.5 &21.5 &28.3 &33.3 &21.0 &26.0 &25.9 &25.2 &22.2 \\
	I-only &27.0 &26.2 &28.4 &24.5 &16.7 &24.7 &26.7 &30.1 &30.1 &25.9 \\
	Q+I  &26.7 &26.2 &26.7 &28.3 &25.0 &21.0 &28.1 &32.2 &21.0 &25.9 \\
	RelNet \cite{bansal2017relnet} & 29.6 &26.2 &34.0 &20.8 &41.7 &29.6 &33.7 &25.2 &28.0 &25.9 \\
	FiLM \cite{perez2018film} &31.7 &28.7 &32.7 &\textbf{39.6} &33.3 &33.3 &29.2 &33.6 &30.8 &29.6\\
	FiLM-BERT \cite{devlin2018bert} & 32.8 &32.9 &33.3 &30.2 &25.0 &32.1 &32.3 &32.2 &34.3 &33.3 \\
	FiLM-BART \cite{lewis2020bart} & 33.0 &32.1 &33.0 &35.8 &50.0 &34.6 &32.6 &37.1 &30.1 &37.0 \\
	\hline
	\textbf{Inter-GPS} (ours) & \textbf{57.5} &\textbf{59.1} &\textbf{61.7} &\text{30.2} &\textbf{50.0} &\textbf{59.3} &\textbf{66.0} &\textbf{52.4} &\textbf{45.5} &\textbf{48.1} \\
	\textbf{Inter-GPS} (GT) & 78.3 &83.1 &77.9 &62.3 &75.0 &86.4 &83.3 &77.6 &61.5 &70.4 \\
	\hline
\end{tabular}
\caption{Evaluation results by our proposed method and compared baselines on the Geometry3K dataset.}
\label{table:result}
\end{table*}

\subsection{Experimental Settings}
\textbf{Datasets and evaluation metrics.}
We conduct experiments on the Geometry3K and GEOS \citep{seo2015solving} datasets. The Geometry3K dataset involves 2,101 training data, 300 validation data, and 601 test data, respectively. The GEOS dataset provides 55 official SAT problems for evaluating geometry solvers. 
Regarding our proposed Inter-GPS model, if the one closest to the found solution among the four choices is exactly the ground truth, the found solution is considered correct.
For a fair comparison, if Inter-GPS fails to output the numeric value of the problem goal within allowed steps, it will randomly choose the one from the four candidates. In terms of compared neural network baselines, the predicted answer has a maximum confidence score among choice candidates.

\begin{table}[th]
\centering 
\small 
\begin{tabular}{{l}|{c}}
	\hline
	\textbf{Method} & Acc (\%)  \\
	\hline
	GEOS \citep{seo2015solving} & 49 \\
	GEOS++ \citep{seo2015solving} & 49 \\
	GEOS-OS \citep{sachan2017learning} & 52 \\
	GEOS++AXIO \citep{sachan2017textbooks} & 55 \\
	\hline
	\textbf{Inter-GPS} (ours)  & \textbf{67}  \\
	\hline
\end{tabular}
\caption{Evaluation results on the GEOS  dataset.}
\label{table:result2}
\end{table}

\noindent\textbf{Baselines.} We implement several deep neural network baselines for geometry solvers to compare them with our method. By default, these baselines formalize the geometry problem solving task as a classification problem, fed by the text embedding from a sequence encoder and the diagram representation from a visual encoder. 
\textit{Q-only} only encodes the problem text in the natural language by a bi-directional Gated Recurrent Unit (Bi-GRU) encoder \citep{Cho2014OnTP}.
\textit{I-only} only encodes the problem diagram by a ResNet-50 encoder \citep{he2016deep} as the input. 
\textit{Q+I} uses Bi-GRU and ResNet-50 to encode the text and diagram, respectively.
\textit{RelNet} \cite{bansal2017relnet} is implemented for embedding the problem text because it is a strong method for modeling entities and relations.
\textit{FiLM} \cite{perez2018film} is compared as it achieves effective visual reasoning for answering questions about abstract images.
\textit{FiLM-BERT}  uses the BERT encoder \cite{devlin2018bert} instead of the GRU encoder, and \textit{FiLM-BART} uses the recently proposed  BART encoder \cite{lewis2020bart}.

\noindent\textbf{Implementation details.} Main hyper-parameters used in the experiments are shown below. For our symbolic solver, a set of 17 geometry theorems is collected to form the knowledge base. For generating positive theorem sequences, each problem is attempted by 100 times with the maximum sequence length of 20. The transformer model used in the theorem predictor has 6 layers, 12 attention heads, and a hidden embedding size of 768. Search steps in Inter-GPS are set up to 100. For the neural solvers, we choose the Adam optimizer and set the learning rate as 0.01, and the maximum epochs are set as 30. Each experiment for Inter-GPS is repeated three times for more precise results.

\subsection{Comparisons with Baselines}

Table \ref{table:result} compares the results of symbolic solver Inter-GPS with baselines on our proposed Geometry3K dataset. Apart from the overall accuracy, the results of different problem types are also reported. Benefiting from symbolic reasoning with theorem knowledge, our Inter-GPS obtains an overall accuracy of 57.5\%, significantly superior to all neural baselines. Inter-GPS even attains a better accuracy compared to human beings. Inter-GPS with ground truth formal language gains a further improvement of 20.8\%. Inter-GPS also obtains state-of-the-art performance over exiting geometry solvers on the GEOS dataset, as shown in Table \ref{table:result2}.

\subsection{Ablation Study and Discussion.} \label{sec:discussion}

\noindent \textbf{Search strategies.} The overall accuracy and average steps needed for solving problems with different search strategies in Inter-GPS are reported in Table \ref{res_table:strategies}. \textit{Predict} refers to the strategy that uses the theorems from the theorem predictor followed by a random theorem sequence. The strategy largely reduces the average steps to 6.5. The final strategy in Inter-GPS applies the predicted theorems first and lower-order theorems in the remain search steps, and gains the best overall accuracy.

\begin{table}[ht]
\centering 
\setlength{\tabcolsep}{3pt}
\small 
\begin{tabular}{l|c|c}
	\hline
	 \textbf{Search strategies} & Accuracy (\%) & \# Steps  \\
	 \hline
	 Random & 75.5 $\pm$ 0.2 & 13.2 $\pm$ 0.1 \\
	 \hline
	 Low-first & 77.3 $\pm$ 0.3 & 15.1 $\pm$ 0.2 \\
	 \hline
	 Predict & 77.5 $\pm$ 0.1 & \textbf{6.5} $\pm$ 0.1 \\
	 \hline
	 Predict+Low-first (final) & \textbf{78.3} $\pm$ 0.1 & 7.1 $\pm$ 0.1 \\
	\hline
\end{tabular}
\caption{Performance of Inter-GPS with different search strategies.}
\label{res_table:strategies}
\end{table}

\noindent \textbf{Problem parsers and literal sources.} The rule-based text parser achieves an accuracy of 97\% while only 67\% for the semantic text parser. Table \ref{res_table:literals} reports the Inter-GPS performance fed with different sources of literals. With literals generated from our problem solver, Inter-GPS achieves an accuracy of 57.5\%.  The current text parser performs very well as there is only a slight gap between Inter-GPS with generated text literals and ground truth literals. An improvement of 17.5\% for Inter-GPS with annotated diagram literals indicates that there is still much space to improve for the diagram parser.

\begin{table}[ht]
\centering 
\setlength{\tabcolsep}{3pt}
\small 
\begin{tabular}{l|c|c|c}
	\hline
	 & Diagram w/o & Diagram & Diagram (GT)  \\
	 \hline
	 Text w/o & 25.0 $\pm$ 0.0 & 46.6 $\pm$ 0.7 & 58.7 $\pm$ 0.2 \\
	 \hline
	 Text  & 25.4 $\pm$ 0.0 & 57.5 $\pm$ 0.2 & 75.0 $\pm$ 0.6 \\
	 \hline
	 Text (GT) & 25.4 $\pm$ 0.0  & 58.0 $\pm$ 1.7 & \textbf{78.3} $\pm$ 0.1 \\
	\hline
\end{tabular}
\caption{Performance of Inter-GPS with predicted and ground truth (GT) literals.}
\label{res_table:literals}
\end{table}

\begin{figure}[ht]
    \centering 
    \includegraphics[width= 0.85\linewidth]{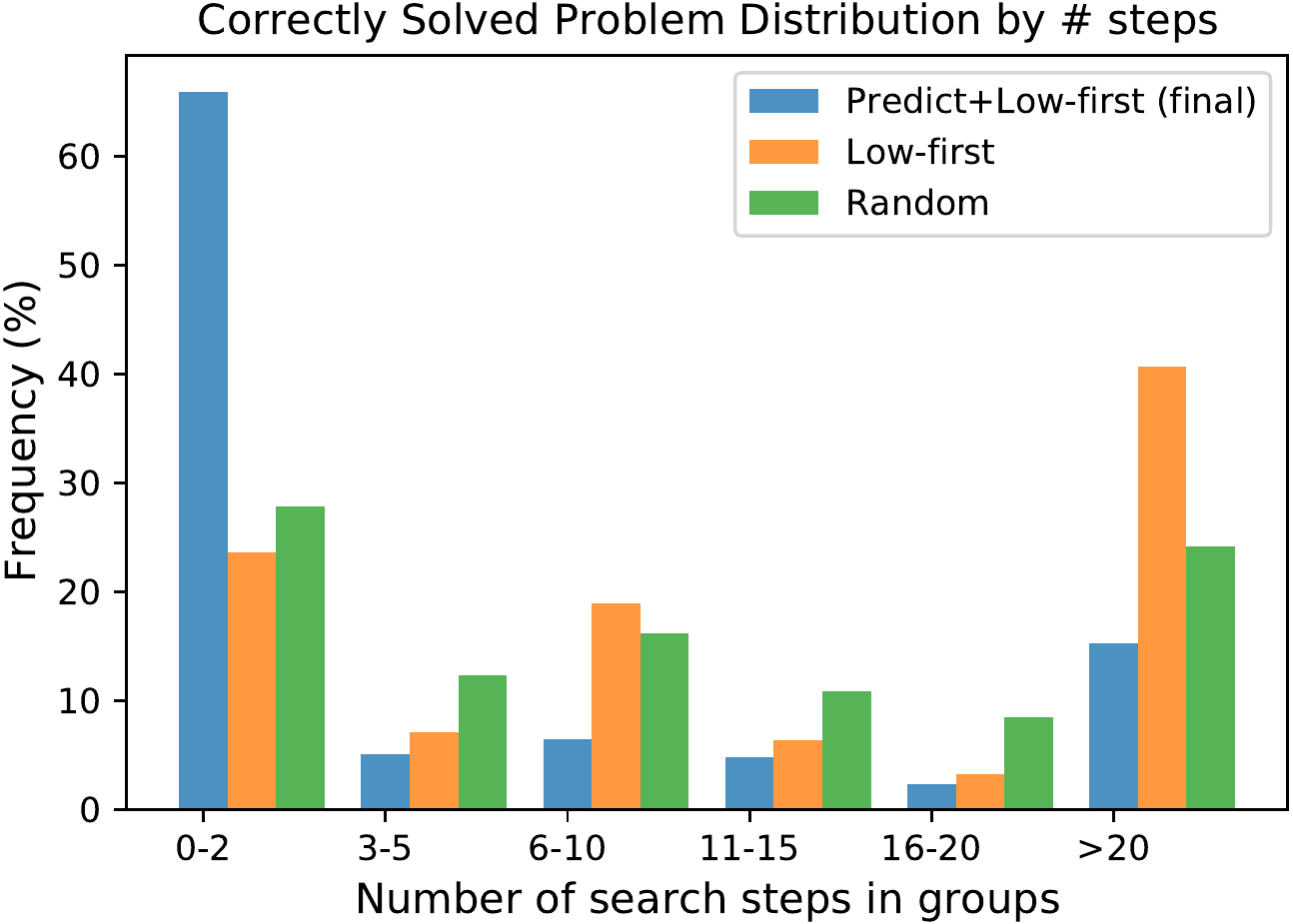}
    \caption{Correctly solved problem distribution by the number of search steps.}
    \label{fig:step_dis}
\end{figure}

\noindent \textbf{Searching step distribution.} Figure \ref{fig:step_dis} compares correctly solved problem distribution by the average number of search steps in different strategies. Our final Inter-GPS applies the \textit{Predict+Low-first} strategy, with which 65.97\% problems are solved in two steps and 70.06\% solved in five steps.

\noindent \textbf{Neural geometry solvers.} Current neural network baselines for geometry solving fail to achieve satisfactory results in the Geometry3K dataset. It is because there are limited data samples for these neural methods to learn meaningful semantics from the problem inputs. Besides, dense implicit representations might not be suitable for logical reasoning tasks like geometry problem solving. We replace the inputs of problem text and diagram in the \textit{Q+I} baseline with the ground truth textual and visual formal annotations and report the result in Table \ref{res_table:nn_solvers}. An improvement of 9.2\% indicates the promising potential for neural network models for problem solving if structural representations with rich semantics are learned.

\begin{table}[ht]
\centering 
\setlength{\tabcolsep}{3pt}
\small 
\begin{tabular}{l|c|c}
	\hline
	 & Diagram (visual) & Diagram (formal)  \\
	 \hline
	 Text (natural) & 26.7 & 35.3  \\
	 \hline
	 Text (formal)  &  34.6 & 35.9 \\
	\hline
\end{tabular}
\caption{Neural solver performance with different representations of the problem text and diagrams.}
\label{res_table:nn_solvers}
\end{table}

\begin{figure}[t]
    \centering 
    \includegraphics[width= 0.93\linewidth]{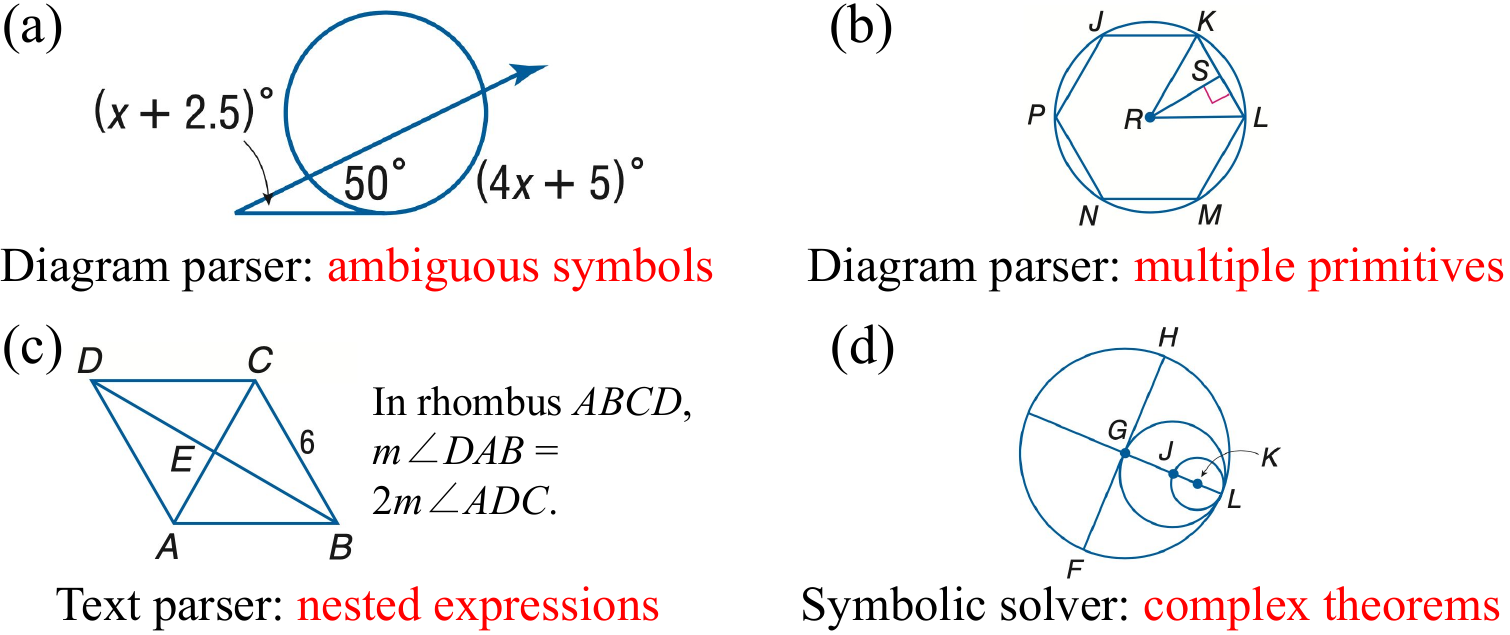}
    \caption{Failure examples for  Inter-GPS.}
    \label{fig:failure}
\end{figure}

\noindent \textbf{Failure cases.} Inter-GPS might not find a solution because of inaccurate parsing results and the incomplete theorem set. Figure \ref{fig:failure} illustrates some failure examples for Inter-GPS. For example, diagram parsing tends to fail if there are ambiguous annotations or multiple primitives in the diagram. It is difficult for the text parser to handle nested expressions and uncertain references. And the symbolic solver is still not capable of solving complex problems with combined shapes and shaded areas in the diagrams.

\noindent \textbf{Interpretability in Inter-GPS.} Inter-GPS provides an interpretable symbolic solver for geometry problem solving. First, Inter-GPS parses the problem contents into a structural representation of formal language. Second, Inter-GPS performs symbolic reasoning to update the geometric relation set explicitly. Last, Inter-GPS applies reasonable theorems sequentially in the search process.

\section{Conclusion}
Solving geometry problems is one of the most challenging tasks in math question answering. In this paper, we propose a large-scale benchmark, Geometry3K, which consists of 3,002 high-school geometry problems with dense descriptions in formal language. We further propose a novel geometry solving approach, \textit{Interpretable  Geometry Problem Solver} (Inter-GPS), which parses the problem as formal language from an automatic parser and performs symbolic reasoning over the theorem knowledge base to infer the answer. Also, a theorem predictor with a low-first search strategy is designed to generate the reasonable theorem application sequence. Experiment results show that Inter-GPS outperforms existing state-of-the-art methods by a large margin. In the future, we plan to extend our work in other math question answering tasks and explore more general symbolic reasoning models.

\section*{Acknowledgments}
This work was supported by MURI N00014-16-1-2007 and DARPA XAI N66001-17-2-4029. We thank Johnson Zhou and Jiahao Li for collecting part of the data. And we thank the help from Jianheng Tang in baseline implementation.

\section*{Ethical Impact}
The problems in Geometry3K are collected from online open sources. The work in this paper may inspire the following research in symbolic reasoning and interpretable models and facilitate education. 
\bibliographystyle{acl_natbib}
\bibliography{anthology,acl2021}

\appendix

\section{Appendix}
We define 91 \textit{predicates} and their corresponding \textit{literal} templates in the geometry language domain. For development, these \textit{predicates} are categorized into six groups: geometric shapes (Table \ref{appex:predicate1}), unary geometric attributes (Table \ref{appex:predicate2}), general geometric attributes (Table \ref{appex:predicate3}), binary geometric relations (Table \ref{appex:predicate4}),  A-IsXOf-B-type geometric relations (Table \ref{appex:predicate5}), as well as numerical attributes and relations (Table \ref{appex:predicate6}). Moreover, \texttt{\$} in the literal templates denotes the undetermined shape.

\begin{table*}[ht]
\centering 
\footnotesize
\begin{tabular}{lll}
	\toprule
	\# & \textbf{Predicates} & \textbf{Literal templates} \\
	\midrule
	1 & \texttt{Point} & \texttt{Point(A)}, \texttt{Point(\$)} \\
	2 & \texttt{Line} & \texttt{Line(A,B)}, \texttt{Line(m)}, \texttt{Line(\$)} \\
	3 & \texttt{Angle} & \texttt{Angle(A,B,C)}, \texttt{Angle(A)}, \texttt{Angle(1)}, \texttt{Angle(\$)} \\
	4 & \texttt{Triangle} & \texttt{Triangle(A,B,C)}, \texttt{Triangle(\$)}, \texttt{Triangle(\$1,\$2,\$3)} \\
	5 & \texttt{Quadrilateral} & \texttt{Quadrilateral(A,B,C,D)}, \texttt{Quadrilateral(1)}, \texttt{Quadrilateral(\$)} \\
	6 & \texttt{Parallelogram} & \texttt{Parallelogram(A,B,C,D)}, \texttt{Parallelogram(1)}, \texttt{Parallelogram(\$)} \\
	7 & \texttt{Square} & \texttt{Square(A,B,C,D)}, \texttt{Square(1)}, \texttt{Square(\$)} \\
	8 & \texttt{Rectangle} & \texttt{Rectangle(A,B,C,D)}, \texttt{Rectangle(1)}, \texttt{Rectangle(\$)} \\
	9 & \texttt{Rhombus} & \texttt{Rhombus(A,B,C,D)}, \texttt{Rhombus(1)}, \texttt{Rhombus(\$)} \\
	10 & \texttt{Trapezoid} & \texttt{Trapezoid(A,B,C,D)}, \texttt{Trapezoid(1)}, \texttt{Trapezoid(\$)} \\
	11 & \texttt{Kite} & \texttt{Kite(A,B,C,D)}, \texttt{Kite(1)}, \texttt{Kite(\$)} \\
	12 & \texttt{Polygon} & \texttt{Polygon(\$)} \\
	13 & \texttt{Pentagon} & \texttt{Pentagon(A,B,C,D,E)}, \texttt{Pentagon(\$)} \\
	14 & \texttt{Hexagon} & \texttt{Hexagon(A,B,C,D,E,F)}, \texttt{Hexagon(\$) }\\
	15 & \texttt{Heptagon} & \texttt{Heptagon(A,B,C,D,E,F,G)}, \texttt{Heptagon(\$)} \\
	16 & \texttt{Octagon} & \texttt{Octagon(A,B,C,D,E,F,G,H)}, \texttt{Octagon(\$)} \\
	17 & \texttt{Circle} & \texttt{Circle(A)}, \texttt{Circle(1)}, \texttt{Circle(\$)} \\
	18 & \texttt{Arc} & 	\texttt{Arc(A,B)}, \texttt{Arc(A,B,C)}, \texttt{Arc(\$)} \\
	19 & \texttt{Sector} & 	\texttt{Sector(O,A,B)}, \texttt{Sector(\$)} \\
	20 & \texttt{Shape} & 	\texttt{Shape(\$)}\\
	\bottomrule
\end{tabular}
\caption{20 predicates and corresponding literal templates for geometric shapes.} 
\label{appex:predicate1}
\end{table*}

\begin{table*}[ht]
\footnotesize
\centering 
\begin{tabular}{lll}
	\toprule
	\# & \textbf{Predicates} & \textbf{Literal templates} \\
	\midrule
	1 & \texttt{RightAngle} & \texttt{RightAngle(Angle(\$))} \\
	2 & \texttt{Right} & \texttt{Right(Triangle(\$))} \\
	3 & \texttt{Isosceles} & \texttt{Isosceles(Polygon(\$))}\\
	4 & \texttt{Equilateral} & \texttt{Equilateral(Polygon(\$))}\\
	5 & \texttt{Regular} & \texttt{Regular(Polygon(\$))}\\
	6 & \texttt{Red} & \texttt{Red(Shape(\$))}\\
	7 & \texttt{Blue} & \texttt{Blue(Shape(\$))}\\
	8 & \texttt{Green} & \texttt{Green(Shape(\$))}\\
	9 & \texttt{Shaded} & \texttt{Shaded(Shape(\$))}\\
	\bottomrule
\end{tabular}
\caption{9 predicates and corresponding literal templates for unary geometric attributes.}
\label{appex:predicate2}
\end{table*}

\begin{table*}[ht]
\footnotesize
\centering 
\begin{tabular}{lll}
	\toprule
	\# & \textbf{Predicates} & \textbf{Literal templates} \\
	\midrule
	1 & \texttt{AreaOf} & \texttt{AreaOf(A)} \\
	2 & \texttt{PerimeterOf} & \texttt{PerimeterOf(A)} \\
	3 & \texttt{RadiusOf} & \texttt{RadiusOf(A)} \\
	4 & \texttt{DiameterOf} & \texttt{DiameterOf(A)} \\
	5 & \texttt{CircumferenceOf}& \texttt{CircumferenceOf(A)}\\
	6 & \texttt{AltitudeOf} & \texttt{AltitudeOf(A)}\\
	7 & \texttt{HypotenuseOf} & \texttt{HypotenuseOf(A)}\\
	8 & \texttt{SideOf} & \texttt{SideOf(A)}\\
	9 & \texttt{WidthOf} & \texttt{WidthOf(A)}\\
	10 & \texttt{HeightOf} & \texttt{HeightOf(A)} \\
	11 & \texttt{LegOf} & \texttt{LegOf(A)} \\
	12 & \texttt{BaseOf} & \texttt{BaseOf(A)}\\
	13 & \texttt{MedianOf} & \texttt{MedianOf(A)}\\
	14 & \texttt{IntersectionOf} & \texttt{IntersectionOf(A,B)}\\
	15 & \texttt{MeasureOf} & \texttt{MeasureOf(A)}\\
	16 & \texttt{LengthOf} & \texttt{LengthOf(A)}\\
	17 & \texttt{ScaleFactorOf} & \texttt{ScaleFactorOf(A,B)}\\
	\bottomrule
\end{tabular}
\caption{17 predicates and corresponding literal templates for general geometric attributes .}
\label{appex:predicate3}
\end{table*}

\begin{table*}[ht]
\footnotesize
\centering 
\begin{tabular}{lll}
	\toprule
	\# & \textbf{Predicates} & \textbf{Literal templates} \\
	\midrule
	1 & \texttt{PointLiesOnLine} & \texttt{PointLiesOnLine(Point(\$),Line(\$1,\$2))}\\
	2 & \texttt{PointLiesOnCircle} & \texttt{PointLiesOnCircle(Point(\$),Circle(\$))}\\
	3 & \texttt{Parallel} & \texttt{Parallel(Line(\$),Line(\$))}\\
	4 & \texttt{Perpendicular} & \texttt{Perpendicular(Line(\$),Line(\$))}\\
	5 & \texttt{IntersectAt} & \texttt{IntersectAt(Line(\$),Line(\$),Line(\$),Point(\$))}\\
	6 & \texttt{BisectsAngle} & \texttt{BisectsAngle(Line(\$),Angle(\$))}\\
	7 & \texttt{Congruent} & \texttt{Congruent(Polygon(\$),Polygon(\$))}\\
	8 & \texttt{Similar} & \texttt{Similar(Polygon(\$),Polygon(\$))} \\
	9 & \texttt{Tangent} & \texttt{Tangent(Line(\$),Circle(\$))}\\
    10 & \texttt{Secant} & \texttt{Secant(Line(\$),Circle(\$))}\\
    11 & \texttt{CircumscribedTo} & \texttt{CircumscribedTo(Shape(\$),Shape(\$))}\\
    12 & \texttt{InscribedIn} & \texttt{InscribedIn(Shape(\$),Shape(\$))}\\
	\bottomrule
\end{tabular}
\caption{12 predicates and corresponding literal templates for binary geometric relations.}
\label{appex:predicate4}
\end{table*}

\begin{table*}[ht]
\footnotesize
\centering 
\begin{tabular}{lll}
	\toprule
	\# & \textbf{Predicates} & \textbf{Literal templates} \\
	\midrule
	1 & \texttt{IsMidpointOf} & \texttt{IsMidpointOf(Point(\$),Line(\$))}\\
	2 &\texttt{IsCentroidOf} & \texttt{IsCentroidOf(Point(\$),Shape(\$))} \\
	3 &\texttt{IsIncenterOf} & \texttt{IsIncenterOf(Point(\$),Shape(\$))}\\
	4 &\texttt{IsRadiusOf} & \texttt{IsRadiusOf(Line(\$),Circle(\$))}\\
	5 & \texttt{IsDiameterOf}& \texttt{IsDiameterOf(Line(\$),Circle(\$))}\\
	6 & \texttt{IsMidsegmentOf} & \texttt{IsMidsegmentOf(Line(\$),Triangle(\$))}\\
	7 &	\texttt{IsChordOf} &\texttt{IsChordOf(Line(\$),Circle(\$))}\\
	8 &	\texttt{IsSideOf} & \texttt{IsSideOf(Line(\$),Polygon(\$))}\\
	9 &\texttt{IsHypotenuseOf} & \texttt{IsHypotenuseOf(Line(\$),Triangle(\$))}\\
    10 & \texttt{IsPerpendicularBisectorOf}	& \texttt{IsPerpendicularBisectorOf(Line(\$),Triangle(\$))}\\
    11 & \texttt{IsAltitudeOf} & \texttt{IsAltitudeOf(Line(\$),Triangle(\$))}\\
    12 & \texttt{IsMedianOf} & \texttt{IsMedianOf(Line(\$),Quadrilateral(\$))}\\
	13 & \texttt{IsBaseOf} & \texttt{IsBaseOf(Line(\$),Quadrilateral(\$))}\\
	14 & \texttt{IsDiagonalOf} & \texttt{IsDiagonalOf(Line(\$),Quadrilateral(\$))}\\
	15 & \texttt{IsLegOf} & \texttt{IsLegOf(Line(\$),Trapezoid(\$))}\\
	\bottomrule
\end{tabular}
\caption{15 predicates and corresponding literal templates for A-IsXOf-B-type geometric relations.}
\label{appex:predicate5}
\end{table*}

\begin{table*}[ht]
\centering 
\footnotesize
\begin{tabular}{lll}
	\toprule
	\# & \textbf{Predicates} & \textbf{Literal templates} \\
	\midrule
	1 & \texttt{SinOf} & \texttt{SinOf(Var)} \\
	2 & \texttt{CosOf} & \texttt{CosOf(Var)} \\
	3 & \texttt{TanOf} & \texttt{TanOf(Var)}\\
	4 & \texttt{CotOf} & \texttt{CotOf(Var)} \\
	5 &	\texttt{HalfOf} & \texttt{HalfOf(Var)}\\
	6 & \texttt{SquareOf} & \texttt{SquareOf(Var)}\\
	7 & \texttt{SqrtOf} & \texttt{SqrtOf(Var)}\\
	8 & \texttt{RatioOf} & \texttt{RatioOf(Var)}, \texttt{RatioOf(Var1,Var2)}\\
	9 &\texttt{SumOf} & \texttt{SumOf(Var1,Var2,...)}\\
	10 & \texttt{AverageOf} & 	\texttt{AverageOf(Var1,Var2,...)}\\
	11 & \texttt{Add} & \texttt{Add(Var1,Var2,...)} \\
	12 & \texttt{Mul} & \texttt{Mul(Var1,Var2,...)}\\
	13 & \texttt{Sub} & \texttt{Sub(Var1,Var2,...)}\\
	14 & \texttt{Div} & \texttt{Div(Var1,Var2,...)}\\
	15 & \texttt{Pow} & \texttt{Pow(Var1,Var2)} \\
	16 & \texttt{Equals} & \texttt{Equals(Var1,Var2)}\\
	17 & \texttt{Find} & \texttt{Find(Var)}\\
	18 & \texttt{UseTheorem} & \texttt{UseTheorem(A\_B\_C)}\\
	\bottomrule
\end{tabular}
\caption{18 predicates and corresponding literal templates for  numerical attributes and relations.}
\label{appex:predicate6}
\end{table*}

\end{document}